\documentclass[letterpaper, 10 pt, journal, twoside]{IEEEtran}

\usepackage{comment}
\usepackage[binary-units]{siunitx}
\usepackage{relsize}
\usepackage{ifthen}
\usepackage[colorinlistoftodos]{todonotes}

\usepackage[vlined,ruled,linesnumbered]{algorithm2e}
\usepackage{graphics} \usepackage{rotating}
\usepackage{color}
\usepackage{enumerate}
\usepackage[T1]{fontenc}
\usepackage{psfrag}
\usepackage{epsfig} \usepackage{booktabs}
\usepackage{graphicx,url}
\usepackage{multirow}
\usepackage{array}
\usepackage{latexsym}
\usepackage{amsfonts}
\usepackage{amsmath}
\usepackage{amssymb}
\usepackage{xstring}
\usepackage[noend]{algorithmic}
\usepackage{multirow}
\usepackage{xcolor}
\usepackage{prettyref}
\usepackage{flexisym}
\usepackage{bigdelim}
\usepackage{breqn} \usepackage{listings}
\let\labelindent\relax
\usepackage{enumitem}
\usepackage{xspace}
\usepackage{bm}
\graphicspath{{./figures/}}
\usepackage{tikz}
\usetikzlibrary{matrix,calc}

\usepackage{mdwlist}

\let\itemize\undefined
\makecompactlist{itemize}{stditemize}

\newrefformat{prob}{Problem\,\ref{#1}}
\newrefformat{def}{Definition\,\ref{#1}}
\newrefformat{sec}{Section\,\ref{#1}}
\newrefformat{sub}{Section\,\ref{#1}}
\newrefformat{prop}{Proposition\,\ref{#1}}
\newrefformat{app}{Appendix\,\ref{#1}}
\newrefformat{alg}{Algorithm\,\ref{#1}}
\newrefformat{cor}{Corollary\,\ref{#1}}
\newrefformat{thm}{Theorem\,\ref{#1}}
\newrefformat{lem}{Lemma\,\ref{#1}}
\newrefformat{fig}{Fig.\,\ref{#1}}
\newrefformat{tab}{Table\,\ref{#1}}

\newcommand{\calE}{{\cal E}}
\newcommand{\calF}{{\cal F}}
\newcommand{\calG}{{\cal G}}
\newcommand{\calH}{{\cal H}}
\newcommand{\calI}{{\cal I}}
\newcommand{\calK}{{\cal K}}
\newcommand{\calL}{{\cal L}}
\newcommand{\calV}{{\cal V}}

\newcommand{\eg}{\emph{e.g.,}\xspace}
\newcommand{\ie}{\emph{i.e.,}\xspace}
\newcommand{\myParagraph}[1]{{\bf #1.}\xspace}

\newcommand{\M}[1]{{\bm #1}} 

\newcommand{\hide}[1]{}

\DeclareMathOperator*{\argmin}{arg\,min}

\newcommand{\blue}[1]{{\color{blue}#1}}

\newcommand{\linkToPdf}[1]{\href{#1}{\blue{(pdf)}}}
\newcommand{\linkToPpt}[1]{\href{#1}{\blue{(ppt)}}}
\newcommand{\linkToCode}[1]{\href{#1}{\blue{(code)}}}
\newcommand{\linkToWeb}[1]{\href{#1}{\blue{(web)}}}
\newcommand{\linkToVideo}[1]{\href{#1}{\blue{(video)}}}
\newcommand{\linkToMedia}[1]{\href{#1}{\blue{(media)}}}

\newcommand{\ph}[1]{\textbf{#1.}}

\def\FINAL{1}

\ifx\FINAL\undefined
\usepackage[paperwidth=10.5in, textheight=9.3in, textwidth=7.14in, left=1.68in, right=1.68in, marginparwidth=1.2in]{geometry}
\fi

\usepackage{times}

\usepackage{multicol}
\usepackage[bookmarks=true,colorlinks=true,urlcolor=red,pagebackref]{hyperref}

\usepackage{flushend}
\usepackage{caption}
\usepackage[font=footnotesize,labelfont=bf]{caption}
\usepackage{subfig}
\usepackage[capitalise]{cleveref}
\usepackage{orcidlink}
\let\oldtexttt\texttt
\renewcommand{\texttt}[1]{{\normalsize\oldtexttt{#1}}}

\renewcommand{\baselinestretch}{0.965}

\ifx\FINAL\undefined
  \newcommand{\wc}[2][]{\todo[color=blue!25, #1]{\textbf{wc}: #2}}
  \newcommand{\wci}[2][]{\textcolor{blue}{\textbf{wc}: #2}}
\else \newcommand{\wc}[2][]{}
  \newcommand{\wci}[2][]{}
\fi

\IEEEoverridecommandlockouts

\begin{document}

\title{
Indoor and Outdoor 3D Scene Graph Generation via Language-Enabled Spatial Ontologies
}

\renewcommand{\linkToPdf}[1]{}\renewcommand{\linkToPpt}[1]{}\renewcommand{\linkToCode}[1]{}\renewcommand{\linkToWeb}[1]{}\renewcommand{\linkToVideo}[1]{}\renewcommand{\linkToMedia}[1]{}

\author{Jared Strader\textsuperscript{\orcidlink{0000-0002-3978-9542}}, Nathan Hughes\textsuperscript{\orcidlink{0000-0002-1201-7032}}, William Chen\textsuperscript{\orcidlink{0009-0002-9193-9197}}, Alberto Speranzon\textsuperscript{\orcidlink{0000-0002-9203-2901}}, and Luca Carlone\textsuperscript{\orcidlink{0000-0003-1884-5397}}\thanks{Manuscript received: December, 8, 2023; Revised February, 15, 2024; Accepted March, 14, 2024.}
\thanks{This paper was recommended for publication by Editor Javier Civera upon evaluation of the Associate Editor and Reviewers' comments.
This work is partially founded by the ARL DCIST program, by the ONR program “Low Cost Autonomous Navigation \& Semantic Mapping in the Littorals”, and by Lockheed Martin Corporation. This research was also partially sponsored by the United States Air Force Research Laboratory and the United States Air Force Artificial Intelligence Accelerator and was accomplished under Cooperative Agreement Number FA8750-19-2-1000. The views and conclusions contained in this document are those of the authors and should not be interpreted as representing the official policies, either expressed or implied, of the United States Air Force or the U.S. Government. The U.S. Government is authorized to reproduce and distribute reprints for Government purposes notwithstanding any copyright notation herein.}
\thanks{J.\,Strader, N.\,Hughes, and L.\,Carlone are with the Laboratory for
Information \& Decision Systems (LIDS), Massachusetts Institute of Technology, Cambridge, MA, USA, {\tt\footnotesize \{jstrader,na26933,lcarlone\}@mit.edu}}\thanks{W.\,Chen is with the Berkeley Artificial Intelligence Research (BAIR), University of California, Berkely, CA, USA, {\tt\footnotesize verityw@berkeley.edu}}\thanks{A.\,Speranzon is with Lockheed Martin,  Advanced Technology Labs, Eagan, MN, USA, {\tt\footnotesize alberto.speranzon@lmco.com}}\thanks{Digital Object Identifier (DOI): see top of this page.}
}

\maketitle

\setcounter{figure}{0}

\begin{abstract}
This paper proposes an approach to build 3D scene graphs in arbitrary indoor and outdoor environments.
Such extension is challenging; the hierarchy of concepts that describe an outdoor environment is more complex than for indoors, and manually defining such hierarchy is time-consuming and does not scale.
Furthermore, the lack of training data prevents the straightforward application of learning-based tools used in indoor settings.
To address these challenges, we propose two novel extensions.
First, we develop methods to build a \emph{spatial ontology} defining concepts and relations relevant for indoor and outdoor robot operation.
In particular, we use a Large Language Model (LLM) to build such an ontology, thus largely reducing the amount of manual effort required.
Second, we leverage the spatial ontology for 3D scene graph construction
using \emph{Logic Tensor Networks} (LTN) to add logical rules, or \emph{axioms} (\eg ``a beach contains sand''), which provide additional supervisory signals at training time thus reducing the need for labelled data, providing better predictions, and even allowing predicting concepts unseen at training time.
We test our approach in a variety of datasets, including indoor, rural, and coastal environments, and show that it leads to a significant increase in the quality of the 3D scene graph generation with sparsely annotated data.
\end{abstract}

\vspace{-2mm}

\IEEEpeerreviewmaketitle

\begin{tikzpicture}[overlay, remember picture]
\path (current page.north east) ++(-4.2,-0.2) node[below left] {
\small This article has been accepted for publication in the IEEE Robotics and Automation Letters.
};
\end{tikzpicture}

\begin{tikzpicture}[overlay, remember picture]
\path (current page.north east) ++(-4.1,-0.6) node[below left] {
\small Please cite as: Jared Strader, Nathan Hughes, William Chen, Alberto Speranzon, Luca Carlone,
};
\end{tikzpicture}
\begin{tikzpicture}[overlay, remember picture]
\path (current page.north east) ++(-4.1,-1) node[below left] {
\small ``Indoor and Outdoor 3D Scene Graph Generation via Language-Enabled Spatial Ontologies'',
};
\end{tikzpicture}
\begin{tikzpicture}[overlay, remember picture]
\path (current page.north east) ++(-7.3,-1.4) node[below left] {
\small in \emph{IEEE Robotics and Automation Letters}, 2024.
};
\end{tikzpicture}

\section{Introduction}

\IEEEPARstart{A}{3D} scene graph~\cite{Armeni19iccv-3DsceneGraphs,Rosinol21ijrr-Kimera, Hughes22rss-hydra,Wu21cvpr-SceneGraphFusion,Hughes24ijrr-hydraFoundations} is a novel map representation that describes the environment as a hierarchical graph grounded in the physical world, where nodes represent spatial concepts and edges represent relationships between concepts.
The last few years have seen significant progress in the development of systems for real-time construction of 3D scene graphs in \emph{indoor} environments.
These approaches typically combine traditional SLAM methods and novel geometric deep learning techniques, such as Graph Neural Networks (GNN) and transformers~\cite{Hughes24ijrr-hydraFoundations}.
The resulting 3D scene graphs include multiple layers describing semantic concepts at different levels of abstraction (\eg from objects to places, rooms, and buildings), which have been proven useful in robotics.
For instance, the benefits of using 3D scene graphs have been demonstrated for path planning~\cite{Rosinol21ijrr-Kimera}, task planning~\cite{Agia22corl-Taskography,Rana23corl-sayplan}, and specific naviation tasks such as object search~\cite{Ravichandran22icra-RLwithSceneGraphs}, by exploiting the semantics and hierarchical structure of the representation.
Although one can envision similar benefits using 3D scene graphs in outdoor environments by exploiting the semantics and hierarchy of the representation~\cite{Berg22icra-outdoorStateAbstractions, Greve23arxiv-urbanDsgs}, no work to date has focused on constructing 3D scene graphs in arbitrary outdoor settings.

\begin{figure}
    \centering
    \includegraphics[width=0.99\columnwidth,trim={0mm 0mm 0mm 0mm},clip]{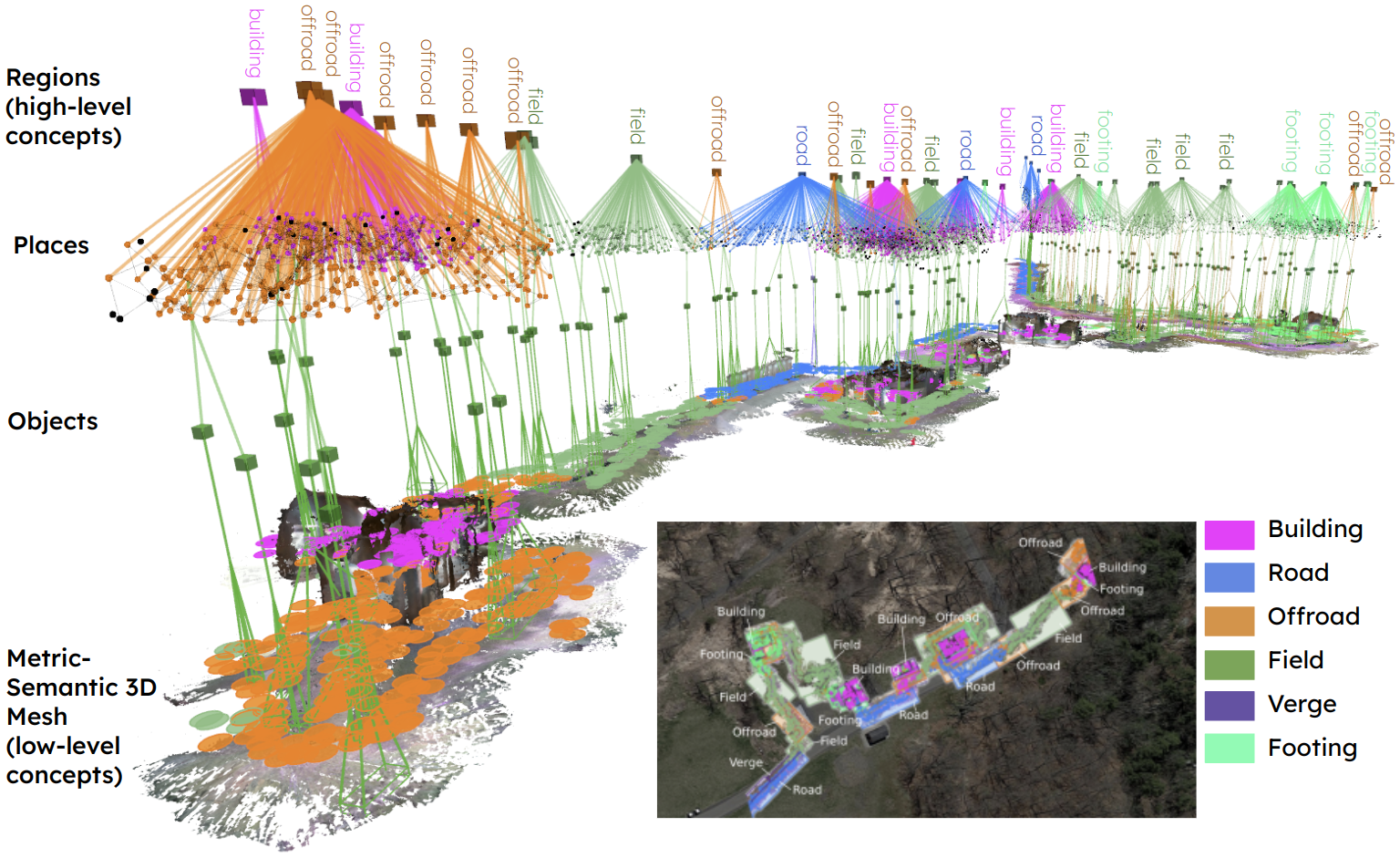}
    \caption{3D scene graph of a indoor-outdoor environment at West Point, NY, constructed using our approach. The hierarchical structure of the graph is based on a spatial ontology generated from Large Language Models, and grounded into a 3D map using Logic Tensor Networks. A satellite image is shown of the area overlaid with the constructed 3D scene graph. \label{fig:coverfigure} \vspace{-7mm} } \end{figure}

Generalizing 3D scene graphs to outdoor environments poses three main challenges:
(i) \emph{Indoor vs. outdoor semantics:} although there is a clear hierarchy of semantic concepts for indoor environments such as objects, rooms, floors, and buildings,
the concepts needed to describe a broad variety of outdoors scenes is less immediate; thus, it is undesirable for a human to manually define such set of labels for each application.
(ii) \emph{Lack of training datasets:} while there are established training datasets one can use for indoor scene graph generation, there is virtually no dataset that allows creating semantically rich 3D scene graphs in outdoor scenes.
For instance, prior work~\cite{Berg22icra-outdoorStateAbstractions} leverages annotations by OpenStreetMap (OSM), but OSM only provides a few annotated classes such as roads, highways, and buildings and does not include smaller objects; this is problematic since existing GNN-based methods~\cite{Hughes24ijrr-hydraFoundations} group nodes (\eg objects) into higher-level concepts (\eg rooms) to build hierarchical representations.
(iii) \emph{Reliability of learning-based methods:} GNN-based approaches can provide incorrect predictions, especially when trained on relatively small datasets or tested outside the training domain; thus, an approach to constrain GNN predictions using common-sense knowledge to improve generalization and accuracy across different types of scenes is desirable.

\myParagraph{Contribution}
We tackle the three challenges above and propose a neuro-symbolic approach to extend 3D scene graph construction to arbitrary environments.
To address the first challenge,
we construct a \emph{spatial ontology} (\cref{sec:ontology}) that describes spatial concepts and relations, which represents common-sense spatial knowledge (\eg a "bathroom" contains a "shower").
While it is well-known that building a comprehensive ontology is difficult and time-consuming, we show that Large Language Models (LLM) can be used to automatically build an ontology,
and we show they can produce a reasonable ontology without fine-tuning and with minimal human effort.
To address the second and third challenge, we resort to a novel learning approach, namely \emph{Logic Tensor Networks} (LTNs)~\cite{Badreddine22ai-LTN} (\cref{sec:ltn}).
In particular, we design a model consisting of a GNN and a multi-layer perceptron (MLP) wrapped inside an LTN~\cite{Badreddine22ai-LTN} that ensures the prediction of the semantic labels in the 3D scene graph satisfy the relations encoded by the spatial ontology.

We evaluate our approach in a variety of datasets, including indoor, rural, and coastal environments (\cref{sec:experiments}).
Our results show that the proposed approach enhances performance with limited training data; for instance, the LTN boosts performance from 12.3\% to 25.1\% for the tested indoor scenes and 29.0\% to 37.2\% (on average) for the tested outdoor scenes when training the proposed model using only 0.1\% of the training data.
We report runtime to demonstrate the proposed approach is suitable for real-time applications (\eg between 0.027s and 0.207s per scene graph in our experiments).
Furthermore, the LTN-based approach predicts labels unseen at training time by leveraging common-sense relations between objects and the location containing them encoded by the spatial ontology.

\section{Related Work}\label{sec:related_work}

\ph{3D Scene Graphs}
The pioneering work of Armeni~\emph{et~al.}~\cite{Armeni19iccv-3DsceneGraphs} introduced the notion of 3D scene graphs as a hierarchical model of 3D environments.
Following this, Rosinol~\emph{et~al.}~\cite{Rosinol21ijrr-Kimera} extended this model to construct 3D scene graphs from sensor data while adding a richer hierarchy of layers.
Other representative approaches also explore constructing 3D scene graphs from geometric data via GNNs~\cite{Wald20cvpr-semanticSceneGraphs} or probabilistic approaches~\cite{Gothoskar21arxiv-3dp3}.
More recently, several approaches explore the creation of 3D scene graphs in real-time in indoor environments~\cite{Wu21cvpr-SceneGraphFusion, Hughes22rss-hydra} or specific outdoor (\ie{} urban) environments~\cite{Greve23arxiv-urbanDsgs}; in contrast, our work focuses on arbitrary indoor and outdoor environments.
Most relevantly, Hughes~\emph{et~al.}~\cite{Hughes24ijrr-hydraFoundations} discuss detecting room categories in 3D scene graphs using a fully-supervised GNN (which requires a significant amount of labeled training data), and Chen~\emph{et~al.}~\cite{Chen22arxiv-LLM2} discuss using large language models for zero-shot room classification in 3D scene graphs.
However, neither of these approaches applies to unstructured (\ie outdoor) environments.
Additionally, 3D scene graphs have started to find many applications in robotics and beyond.
Recent works have successfully used 3D scene graphs to exploit the semantics and hierarchical structure for navigation~\cite{Ravichandran22icra-RLwithSceneGraphs,Seymour22icpr-graphMapper} as well as task and motion planning~\cite{Agia22corl-Taskography, Rana23corl-sayplan}.
Finally, 3D scene graphs also have nascent applications outside of robotics~\cite{Dhamo21iccv-dsgSceneGeneration,Tahara20ismar-retargetableAR}.

\ph{Ontologies}
To build and share a meaningful representation of a 3D environment, a robot needs a common "vocabulary" describing the concepts and relations observed by the robot.
Such a vocabulary may be represented as an \emph{ontology}.
The precise definition of ontology varies between communities~\cite{Gruber95ijhcs-DesignOfOntologies},
and the term is often used interchangeably with \emph{knowledge graph} and \emph{knowledge base}.
Typically, an ontology refers to the ``schema'' of data (\ie abstract concepts and relations) whereas a knowledge graph is an instance of an ontology populated with data.
We adopt the definition of knowledge base from~\cite{Badreddine22ai-LTN} (more details in~\cref{sec:ltn}). A significant effort has been made in the creation of common-sense ontologies and knowledge graphs~\cite{Auer07sw-DBpedia, Speer17aaai-ConceptNet}
, and there has been a surge applying these to problems such as 2D scene graph generation~\cite{Zareian20eccv-BridgingKnowledgeSGG}, image classification~\cite{Movshovitz15cvpr-OntoStoreClass}, and visual question answering~\cite{Zheng21pr-EmbeddingDesignVQA}, to name a few.
Most relevant to our work, Qiu~\emph{et~al.}~\cite{Qiu23arxiv-knowledgeGraphSSG} and Zhang~\emph{et~al.}~\cite{Zhang21neurips-knowledgeInspiredSceneGraphs} use ontologies to inform the generation of indoor 3D scene graphs, and Hsu~\emph{et~al.}~\cite{Hsu23cvpr-ns3d} discuss how to use LLMs to generate grounding predicates for object and their relationships.
Constructing common-sense ontologies and knowledge graphs is an open problem; many authors have recently explored the role and use of LLMs to generate these representations.
These works discuss either how to directly create common-sense representations by refining the LLM output~\cite{Chen21acl-constructingTaxonomies} or evaluate the common-sense knowledge accuracy of LLMs~\cite{Sun23arxiv-headToTail}.
The interested reader is referred to~\cite{Pan23arxiv-llmOpportunities} for a discussion of challenges in this area.

\section{Problem Statement}
\label{sec:problem}
A 3D scene graph (\eg \cref{fig:coverfigure}) is a hierarchical graph~\cite{Hughes24ijrr-hydraFoundations} where nodes and edges represent spatial concepts and relations grounded in the physical world, and are grouped into \emph{layers} corresponding to different levels of abstraction of the scene.
In indoor environments, the layers from bottom to top consist of a \emph{metric-semantic 3D mesh}, \emph{objects and agents}, \emph{places} (a topological representation of free space), and \emph{rooms} (groupings of places describing rooms).
While existing work on 3D scene graphs focus on grounding concepts for indoor environments, we extend the 3D scene graph model in~\cite{Rosinol21ijrr-Kimera, Hughes22rss-hydra} to ground concepts in arbitrary environments (\eg{} indoor and outdoor).
Specifically, given the low-level layers (\ie{} metric-semantic 3D mesh, objects and agents, and places) which have the same meaning in both indoor and outdoor environments and can be built as in~\cite{Hughes22rss-hydra}, we construct the high-level layers (\ie regions layer generalizing the rooms layer) of a 3D scene graph based on a given set of spatial concepts (\eg{} ``kitchen'' and "bathroom" for indoors or ``road'' and ``field'' for outdoors).

\section{Language-enabled Spatial Ontologies}
\label{sec:ontology}

This section introduces the notion of \emph{spatial ontology} and describes how to generate one using LLMs using two different approaches: text scoring and text completion.

\begin{figure}
    \centering
    \includegraphics[width=0.99\columnwidth]{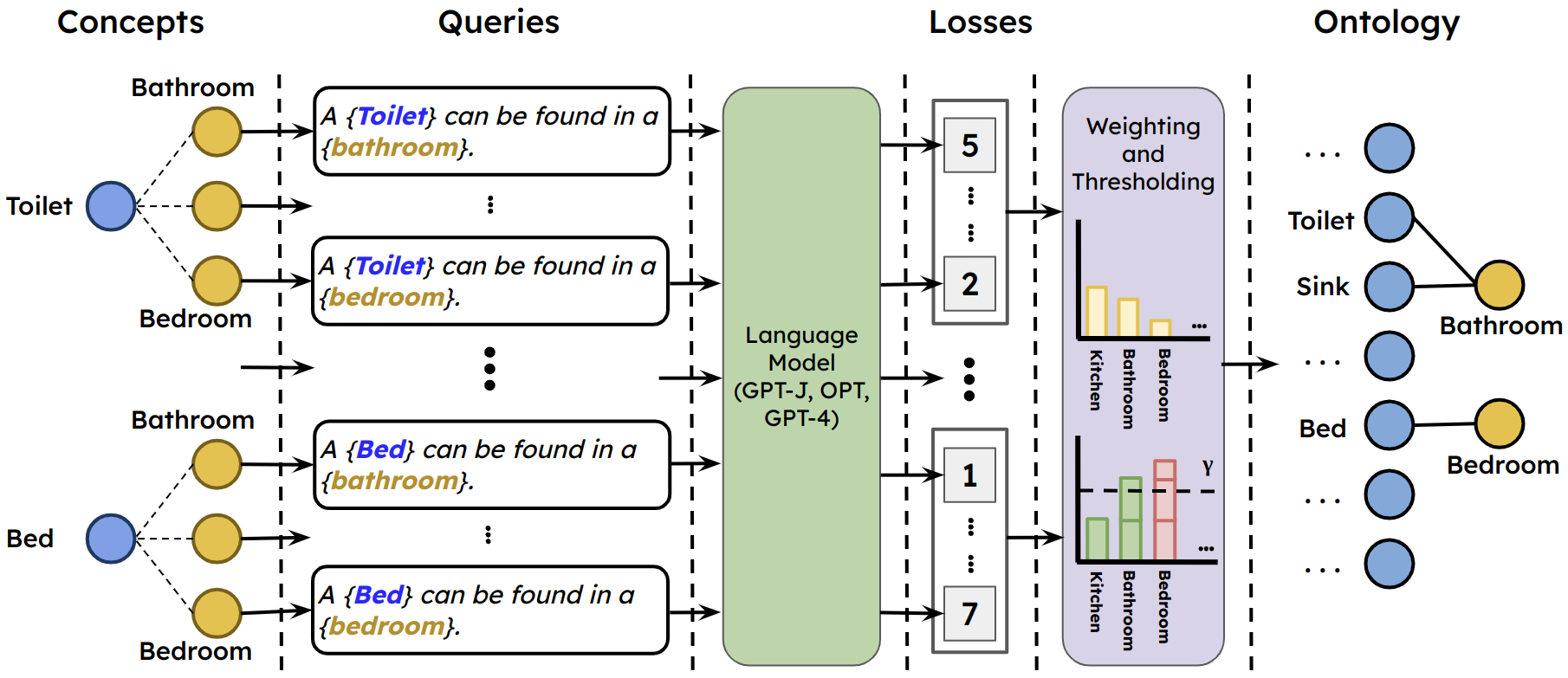} \caption{
        Text scoring approach for generating spatial ontology. The language model assigns a loss to the text string between low-level and high-level labels. The loss is rescaled using a softmax to assign weights to the edges, then the lowest weighted edges are pruned away.
        \vspace{-4mm} \label{fig:text_scoring}}
\end{figure}

\begin{figure}
    \centering
    \includegraphics[width=0.99\columnwidth]{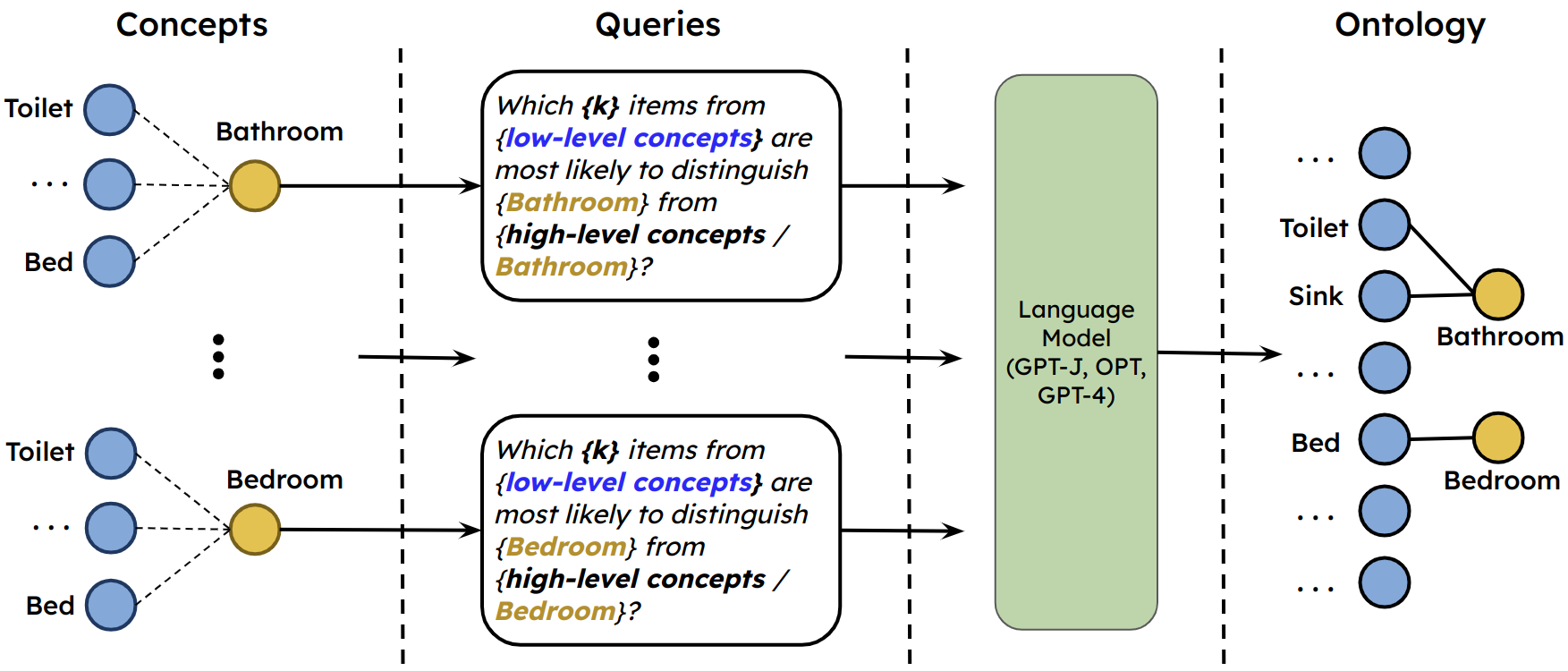} \caption{
        Text completion approach for generating spatial ontology. The language model is queried asking what low-level labels distinguish high-level labels. The response is parsed to generate the spatial ontology.
        } \vspace{-6mm} \label{fig:text_completion}
\end{figure}

\subsection{Spatial Ontology}
A spatial ontology is a representation of spatial knowledge, which can be modeled as a graph where nodes are (abstract) spatial concepts and edges are spatial relations between the concepts.
In this work, we consider a spatial ontology where each node corresponds to either low-level spatial concepts $\calL = \{\ell_1, \cdots, \ell_{n}\}$ or high-level spatial concepts $\calH = \{ h_1, \cdots, h_{m}\}$ where the $n$ low-level and $m$ high-level concepts correspond to semantics captured by the mesh layer (\eg sand, rock) and locations captured by the places layer (\eg beach), respectively.
Furthermore, we only consider \emph{inclusion} relations, which describe where low-level concepts may be located with respect to high-level concepts. For example, an edge between a low-level concept $\ell \in \calL$ (\eg{} ``sink'') and a high-level concept $h \in \calH$ (\eg{} ``kitchen'') means that $\ell$ is located within $h$ (\eg{} a ``sink'' is located in a ``kitchen'').
This implies that the spatial ontology is a bipartite graph with $|\calH| \times |\calL|$ biadjacency matrix $\M{\Omega}$.\footnote{The adjacency matrix for a bipartite graph has the form ${\begin{bmatrix} \mathbf{0} & \M{\Omega} \\ \M{\Omega}^{\mathsf{T}} & \mathbf{0} \end{bmatrix} }$ where $\M{\Omega}$ is the biadjacency matrix. In our problem, the structure of the adjacency matrix formalizes the fact that the edges only relate low-level concepts (\eg objects) with high-level concepts (\eg locations).}
In contrast to a 3D scene graph, which is a \emph{grounded} representation of the environment, a spatial ontology is independent from the specific environment observed by the robot.
For example, the concept of ``chair'' in an ontology is not associated with a particular instance of a ``chair'', while in a 3D scene graph, the concept of ``chair'' is grounded in the scene the robot observes.
In other words, an ontology is a graph representing all possible concepts and their relations (\ie it describes that chairs are typically in kitchens), while a 3D scene graph is a model of a specific environment (\ie{} an instantiation of a spatial ontology with data).

\subsection{Inferring Spatial Ontologies from LLMs}
A spatial ontology may be defined manually by an expert, but such manual effort is time-consuming and does not scale well to large ontologies.
While there is a history of using LLMs to evaluate relations between concepts, we focus on directly creating a spatial ontology, which can be leveraged for constructing 3D scene graphs.
We describe two approaches (text scoring and text completion) to use LLMs to build spatial ontologies at scale, and while text completion is the preferred method used in our approach, we include text scoring for completeness since both generate reasonable ontologies based on our experiments (more in~\cref{sec:experiments}).

\subsubsection{Text Scoring} \label{sec:text_scoring}
In this approach, we initially assume edges exist between all pairs of low-level and high-level concepts.
Weights are assigned to each edge in the ontology each representing if a low-level concept is located within a high-level concept.
The weights on the edges are assigned using an auto-regressive language model $\Lambda$, which is trained to assign probabilities to text strings.
More precisely, $\Lambda$ maps strings of tokens $W$ to the log probability of the string written as $\Lambda(W) \approx \text{log} \, p(W)$.
The template $g_{score}$ for generating the text string between the $i$th low-level concept $\ell_i \in \calL$ and the $j$th high-level concept $h_j \in \calH$ is given by
\begin{equation}
    g_{score}(i,j) = \text{``} \ell_i \; \text{is often found in} \; h_j \text{''}. \nonumber
\end{equation}
The language model assigns a loss to each edge based on the associated text string, which corresponds to the log probability the text string will occur given the model parameters, then the weights for each edge are rescaled using a softmax function:
\begin{equation}
    w_{ij} = \frac{\exp \{ \Lambda(g_{score}(i,j))/K\}}{\sum_{k=1}^{|H|}\exp \{ \Lambda(g_{score}(i,k))/K \}},
\end{equation}
where $K \in (0,\infty]$ is a temperature parameter.
The weights for the edges incident to each low-level concept are accumulated in descending order by weight until exceeding a threshold $\gamma$, then the edges not reached are pruned from the spatial ontology; formally, the $r$ retained edges for the $i$th low-level concept satisfy $\sum_{j=1}^{r-1} w_{ij} \leq \gamma$ and $\sum_{j=1}^{r} w_{ij} > \gamma$ where $\gamma$ is a user-defined threshold. \Cref{fig:text_scoring} provides an overview of the overall approach.

\subsubsection{Text Completion} \label{sec:text_completion}
While the approach described above generates reasonable ontologies (more in~\cref{sec:experiments}), the result is an ontology with edges for the most likely relations (\eg a ``wall'' is more likely to be located in a ``bathroom'' than a ``shower''); instead, the ontology should include edges that are useful for distinguishing between different high-level concepts based on low-level concepts (\eg a ``shower'' is more likely to distinguish a ``bathroom'' from a ``bedroom'' than a ``wall'').
Here, we construct the spatial ontology by querying the language model for each high-level concept to generate the $k$ edges connecting low-level concepts to each high-level concept.
The template $g_{comp}$ for generating the text string for the $k$ edges associated with the $i$th high-level concept $h_i \in \calH$ is given by
\begin{align}
    g_{comp}&(i,k) = \text{``Which } \{k\} \text{ items from } \{\calL\} \text{ are most } \nonumber \\[-0.65ex]
    & \text{likely to distinguish } \{h_i\} \text{ from } \{\calH / h_i\}. \text{ Answer} \nonumber \\[-0.65ex]
    & \text{with a python list using exact strings in }  \{\calL\} \text{.''} \nonumber
\end{align}
After querying the language model, the response requires post-processing since the language model may not format the response as a python list or hallucinate concepts (\ie concepts not from $\calL$).
If the response is improperly formatted, the formatting is corrected in post-processing, and if hallucinated concepts are in the response, the text string used for the query is appended with ``Do not respond with concepts in $\{\calL'\}$'' where $\calL'$ is the set of concepts returned during this procedure not in the given set of concepts $\calL$.
The response generated by the language model is stochastic; thus, each query is repeated $N$ times, and the top $k$ edges generated are used in the spatial ontology; see \cref{fig:text_completion}.

\section{Leveraging Spatial Ontologies \\ for 3D Scene Graph Generation}\label{sec:ltn}
This section briefly reviews LTNs and shows how to generate 3D scene graphs using LTNs.
Specifically, we show how to build a knowledge base from a spatial ontology, which is leveraged for constructing 3D scene graphs.

\subsection{Logic Tensor Networks}
A Logic Tensor Network (LTN)~\cite{Badreddine22ai-LTN} is a neuro-symbolic framework that combines explicit symbolic knowledge expressed as a set of logical formulas with implicit subsymbolic knowledge (e.g., encoded within the parameters of a neural network).
LTNs support tasks such as learning (making generalization from data), reasoning (verifying a consequence from facts), and query-answering (evaluating the truth value of a logical expression).
To express the relationship between symbolic and subsymbolic knowledge, LTNs are built on Real Logic~\cite{Fagin20arxiv-realLogic}, which is a differentiable first-order language with truth values in $[0,1]$.
This allows converting Real Logic formulas (\eg{} a differentiable loss) into computational graphs that enable gradient-based optimization.

\subsubsection{Syntax of Real Logic}
Formally, Real Logic is defined on a first-order language $\calF$ with a signature\footnote{The signature of a language is the set of symbols with meanings that vary with the interpretation (\ie non-logical symbols) and exclude connectives and quantifiers (\ie logical symbols).} consisting of sets of symbols for constants, variables, functions, and predicates.
A formula (or \emph{axiom})  in $\calF$ is an expression constructed using symbols in the signature together with logical connectives, $\neg, \land, \lor, \rightarrow$, and quantifiers, $\forall, \exists$, following the same syntax\footnote{The syntax is the set of rules for constructing well-formed expressions.} as First-Order Logic (FOL).
For example, we can construct a formula stating that ``every \emph{kitchen} contains a \emph{sink}'' as $\forall y (\texttt{Kitchen}(y) \rightarrow \exists x (\texttt{Sink}(x) \land \texttt{PartOf}(x,y))$, where $x,y$ are variables and $\texttt{Sink}(\cdot)$, $\texttt{Kitchen}(\cdot)$, $\texttt{PartOf}(\cdot,\cdot)$ are predicates.
In real-world scenarios, a formula may be partially true (\eg a \emph{kitchen} may exist that does not contain a \emph{sink}); thus, Real Logic adopts continuous truth values in $[0,1]$.

\subsubsection{Semantics of Real Logic}
The semantics of Real Logic are the rules for interpreting symbols with respect to the real world, thus explicitly connecting symbolic and subsymbolic knowledge.
Intuitively, interpretations\footnote{In~\cite{Badreddine22ai-LTN}, the authors use the terminology \emph{grounding} instead of \emph{interpretation}, the latter is standard in mathematical logic and preferred here.} $\calI$ are the numeric values associated with each symbol, as well as
how to numerically implement the operations applied to such values.
In general, constants and variables are interpreted as tensors, whereas
operations include \emph{functions}, which map tensors to tensors, and
\emph{predicates}, which map tensors to real numbers in $[0,1]$.
For example, a variable $x$ may be interpreted as a feature vector associated to an object observed by the robot (\eg \emph{sink}), and the predicate $\texttt{Sink}(\cdot)$ may be interpreted as a binary classifier that takes a feature vector as input and returns if the input is a \emph{sink} or not (\eg returning a truth value between 0 and 1). In Real Logic, a formula $\phi$ under an interpretation $\calI$ is satisfied if $\calI(\phi) = 1$, but in practice, a formula $\phi$ is typically only partially satisfied (\ie $\calI(\phi) \in [0,1]$).
Since predicates are partially true or false, there are various ways of defining connectives and quantifiers (\eg{} see ~\cite{Van22ai-diffFuzzyLogic}).\footnote{For example, in Łukasiewicz logic~\cite{Van22ai-diffFuzzyLogic}, $\calI(\phi_1 \land \phi_2) = \max(0, \calI(\phi_1)+\calI(\phi_2)-1)$ and $\calI(\phi_1 \lor \phi_2) = \max(1, \calI(\phi_1)+\calI(\phi_2))$ where $\phi_1$ and $\phi_2$ are formulas.}

\subsubsection{Learning with LTNs}
If we are given a set of formulas in Real Logic, LTNs allow learning the interpretation of symbols in the signature.
For example, if constants are interpreted as embeddings, and functions and predicates are interpreted as neural networks, the interpretation $\calI$ depends on learnable parameters $\theta$ with a parametric interpretation $\calI(\cdot|\theta)$.
The symbols in the signature, logical formulas, and interpretations together are referred to as a \emph{Real Logic knowledge base}.
The optimization problem underlying LTNs for learning interpretations is a \emph{satisfiability problem}, which yields the values of the parameters $\theta^*$ that maximize the truth values of the conjunction of a set of logical formulas $\calK$ in a Real Logic knowledge base.
We can write this as:
\begin{equation}
\label{eq:sat}
\theta^* = \argmin_{\theta \in \Theta} \left( 1 - \underset{\phi \in \calK}{\texttt{SatAgg}} \, \calI(\phi|\theta) \right)
\end{equation}
where $\texttt{SatAgg}: [0,1]^* \rightarrow [0,1]$ is an aggregating operator over all the formulas $\phi \in \calK$, typically implemented numerically as a generalized mean~\cite{Badreddine22ai-LTN}.
Since Real Logic interprets expressions as real-valued tensors, LTNs associate gradients to the expressions in the knowledge base and optimizes~\eqref{eq:sat} via gradient-based optimization.

\begin{figure}[t]

\subfloat[Learning Phase. In the learning phase, the place features are processed by the GNN to compute node embeddings, and the MLP processes the embeddings and returns the softmax of the logits. The MLP is wrapped in an LTN to compute the satisfaction of a Real Logic knowledge base using the data-driven axioms based on ground truth labels and ontology-driven axioms based on the aggregated semantics from the mesh layer.]{\noindent\includegraphics[width=0.99\columnwidth]{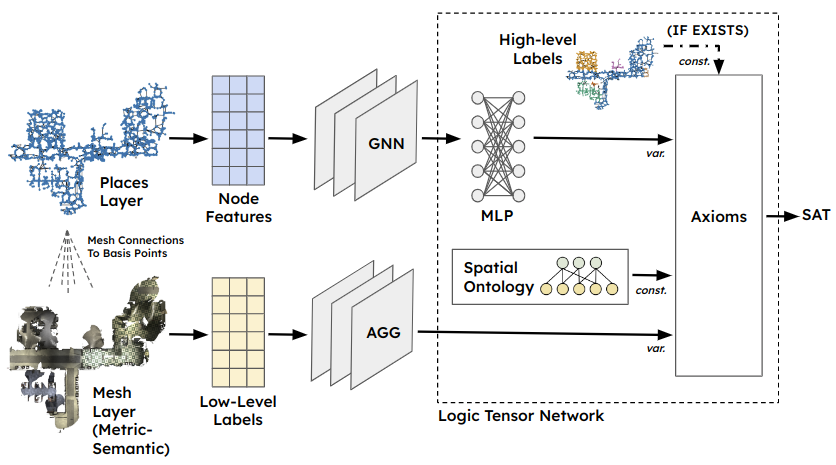}
}

\subfloat[Inference Phase. In the inference phase, the LTN is dropped, and the model predicts the node labels based on the place features using the trained MLP and GNN from the learning phase.]{\noindent\includegraphics[width=0.99\columnwidth]{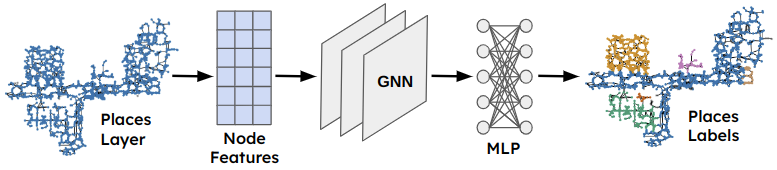}
}

\caption{Proposed architecture for learning and inference.\label{fig:arch} }
\vspace{-7mm}

\end{figure}
\subsection{Formalizing 3D Scene Graph Generation in LTNs}
This section presents our approach to create hierarchical 3D scene graphs with a spatial ontology using LTNs.
Specifically, given the low-level layers of a 3D scene graph (\ie metric-semantic 3D mesh, objects and agents, and places), we construct the high-level layers based on a given spatial ontology and labeled examples (if available).
We formulate this as a node classification problem on the places layer using a GNN for generating node embeddings and an MLP for predicting node labels.
We leverage LTNs during training to constrain the model to make predictions consistent with the spatial ontology.
The proposed architecture for learning and inference is given in~\cref{fig:arch}.
We train the model end-to-end to learn the parameters of both the GNN and MLP based on the satisfaction of a knowledge base, as described below.

\subsubsection{Constants and Variables}
While constants and variables are used in the same way in functions and predicates, the difference is that variables can appear in quantifiers (\ie a variable can have different instances with different values while constants are always a fixed value).

\ph{Constants}
The graph representing the places layer of a 3D scene graph is $\calG = (\calV, \calE)$ where $\calV$ is the set of place nodes and $\calE$ is the set of edges between place nodes.
The label space is $\{y_1,\cdots,y_m\}$, which corresponds to the $m$ high-level concepts $\calH$ in the spatial ontology; specifically, $y_i$ is the one-hot vector for $h_i \in \calH$.
The spatial ontology is a constant $\M{\Omega}$ interpreted as the $|\calH| \times |\calL|$ biadjacency matrix (\eg built as described in~\Cref{sec:ontology}), which defines the valid choices of a high-level concepts for a place node given the low-level concepts observed by the robot.

\ph{Variables}
The GNN associates an embedding $x_v$ to each place node $v \in \calV$.
This embedding is passed to an MLP to classify the high-level concept (\ie region) the node belongs to.
At the same time, we attach a second (pre-computed) feature vector $q_v$ to each place node $v \in \calV$, describing the
low-level concepts observed by the robot (\eg nearby objects).
More in detail,
each place node is connected to nearby nodes in the metric-semantic 3D mesh, called basis points (see~\cite{Hughes24ijrr-hydraFoundations}); the semantic labels of the basis points are one-hot vectors representing the low-level concepts $\calL$ in the ontology.
We build $q_v$ as the histogram of semantic labels of the basis points connected to a node $v \in \calV$; the low-level concepts are captured by the basis points; thus, the object and agent layers are not required.

\subsubsection{Predicates}
The predicates take constants and variables as input and return a real number in $[0,1]$.
Here, we propose predicates, which are either data-driven (referred to as \emph{equivalence} predicates) or ontology-driven (referred to as \emph{inclusion} predicates).
This allows learning the interpretations of predicates parameterized by the parameters $\theta$ of our model using a combination of labeled data and unlabeled data.
The equivalence predicates rely only on the constants $y_i$ (\ie the labeled data) while inclusion predicates rely only on the constant $\M{\Omega}$ (\ie only use the spatial ontology).

\ph{Equivalence}
We propose $\texttt{IsClassOf}(\cdot,\cdot|\theta)$ as an equivalence predicate, which quantifies if a high-level concept predicted by the $\texttt{MLP}_\theta(\cdot)$ for a place node matches the ground truth label (if available at training time):
\begin{equation}
    \calI( \texttt{IsClassOf}|\theta): x_v^{(i)}, y_i \mapsto y_i^T \cdot \texttt{MLP}_\theta(x_v^{(i)}) \;\;
\end{equation}
where $x_v^{(i)}$ denotes the embedding for a place node $v \in \calV$ with label $y_i$.
In words, the predicate measures the agreement between the predicted label $\texttt{MLP}_\theta(x_v^{(i)})$ and the
ground truth label $y_i$, and represents the statement: ``A place node with label $y_i$ is classified as $y_i$''.
Intuitively, this predicate plays a similar role as cross-entropy loss in supervised learning.

\ph{Inclusion}
At training time, we encourage the model (\ie GNN+MLP) to predict labels for place nodes that match the spatial ontology based on the connected basis points.
More formally, we build a formula stating ``If the spatial ontology has connections valid with the basis points connected to a place node, the place node is classified with a label similar to the ones encoded in the spatial ontology''.
The first part of the formula is to circumvent missing data in the spatial ontology: for instance, the ontology may have disconnected low-level concepts,
and we do not use the ontology in those cases.
The second part encourages the prediction to agree with at least one of the edges in the spatial ontology.
In order to formalize this formula, we need two predicates: $\texttt{IsValid}(\cdot,\cdot)$ and $\texttt{IsSimilar}(\cdot,\cdot,\cdot|\theta)$, which we describe below.

In order to assert the validity of the low-level concepts $q_v$ attached to a node $v \in \calV$ (\ie that low-level concepts connected to a place node are connected to high-level concepts in the ontology),
we define a predicate $\texttt{IsValid}(\cdot,\cdot)$:
\begin{align}
    \calI(\texttt{IsValid}): \M{\Omega},q_v \mapsto \texttt{sum}(\hat{\M{\Omega}} \cdot \hat{q_v})
\end{align}
where $\hat{\cdot}$ is an operator normalizing columns with respect to the $\ell_1$-norm.
Intuitively, $\texttt{IsValid}(\M{\Omega},q_v)=0$ if none of the low-level concepts in the histogram $q_v$ are connected to a high-level concept in the spatial ontology $\M{\Omega}$.

Next, we define the predicate $\texttt{IsSimilar}(\cdot,\cdot,\cdot|\theta)$ to quantify the degree that high-level concepts predicted by the MLP are similar to the expected high-level concepts in the spatial ontology based on the low-level features $q_v$:
\begin{align}
    \calI(\texttt{IsSimilar}|\theta): x_v,\M{\Omega},q_v \mapsto \texttt{MLP}_\theta(x_v)^T \cdot \hat{\M{\Omega}} \cdot \hat{q_v}.
\end{align}
Intuitively, $\hat{\M{\Omega}} \cdot \hat{q_v}$ represents the valid high-level concepts associated to low-level concepts in $q_v$ (according to the ontology $\M{\Omega}$), and the predicate computes the similarity between these high-level concepts and the ones predicted by the model, which enforces alignment between the model predictions $\texttt{MLP}_\theta(x_v)$ and ontology predictions $\hat{\M{\Omega}} \cdot \hat{q_v}$ .

\subsubsection{Axioms}
The axioms are the logical formulas constructed from constants, variables, and predicates representing statements.
We distinguish the axioms by whether they use the equivalence or the inclusion predicates.
The equivalence axioms $\phi_{\text{Equiv}}$ and inclusion axioms $\phi_{\text{Incl}}$ are:
\begin{align}
    \phi_{\text{Equiv}} \!:& \; \forall  x_v^{(i)} \; \texttt{IsClassOf}(x_v^{(i)},y_i) \nonumber \\
    \phi_{\text{Incl}} \!:&\; \forall  x_v, q_v \;\texttt{IsValid}(\M{\Omega},q_v) \! \rightarrow \texttt{IsSimilar}(x_v,\M{\Omega}, q_v) \nonumber
\end{align}
where diagonal quantification is used for both $\phi_{\text{Equiv}}$ and $\phi_{\text{Incl}}$ as described in~\cite{Badreddine22ai-LTN}.
Here, $x_v^{(i)}$ corresponds to place nodes with label $y_i$, and
 each place node $v \in \calV$ is assigned features $x_v$ and $q_v$; thus, we only want to construct statements between $x_v$ and $q_v$ for each $v \in \calV$.
At training time, the equivalence axioms promote the model to make predictions that agree with the annotated labels (if available), while the inclusion axioms encourage the model to make prediction that agree with the spatial ontology based on the semantics of the connected basis points.
In general, the equivalence axioms may contradict the inclusive axioms if the annotated labels disagree with the spatial ontology; in such cases, both types of axioms will influence the predictions.

\begin{figure*}
\subfloat[MP3D\label{fig:mp3d_results}]{\centering
        \includegraphics[width=0.32\textwidth]{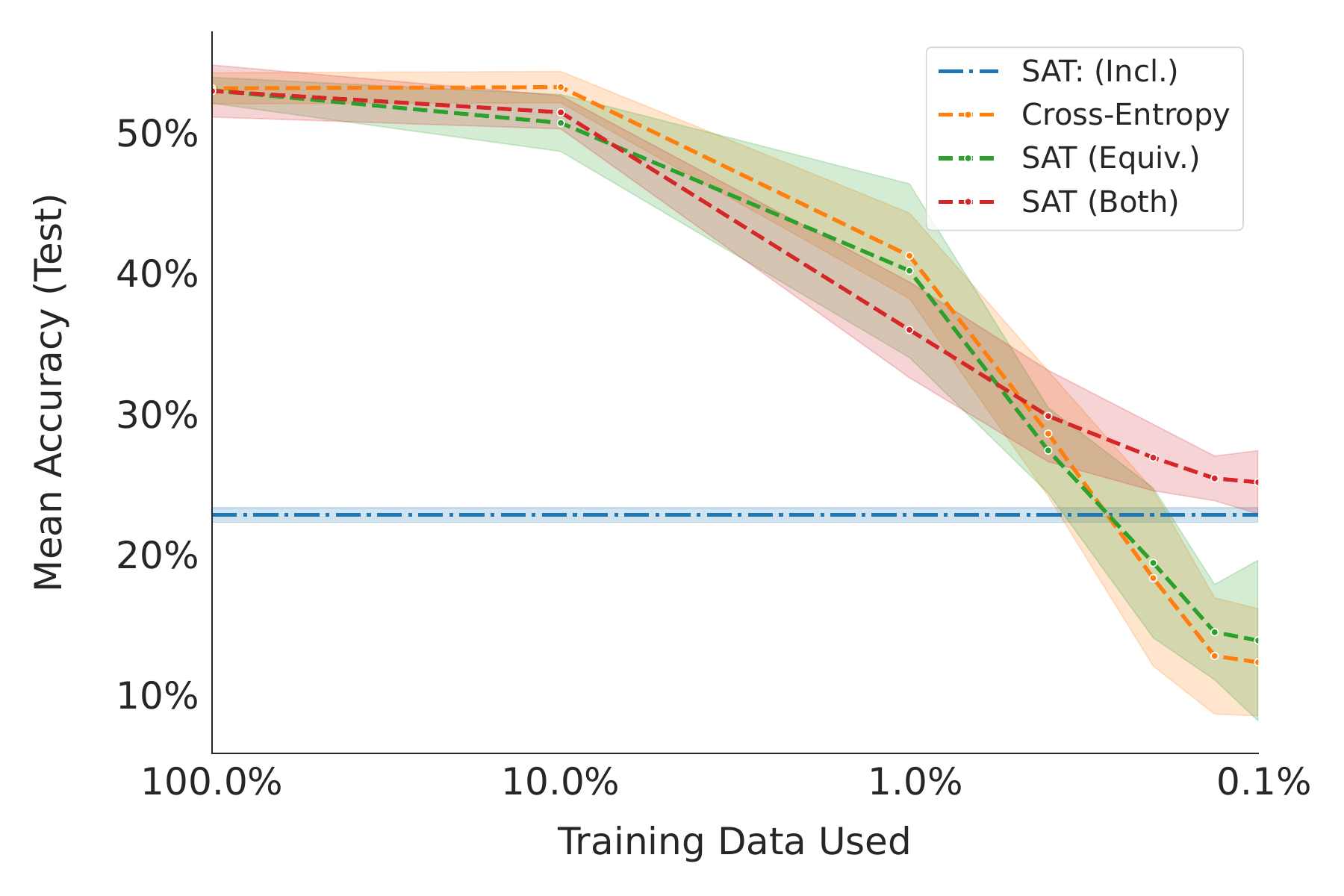}
    }
    \subfloat[West Point\label{fig:westpoint_results}]{\centering
        \includegraphics[width=0.32\textwidth]{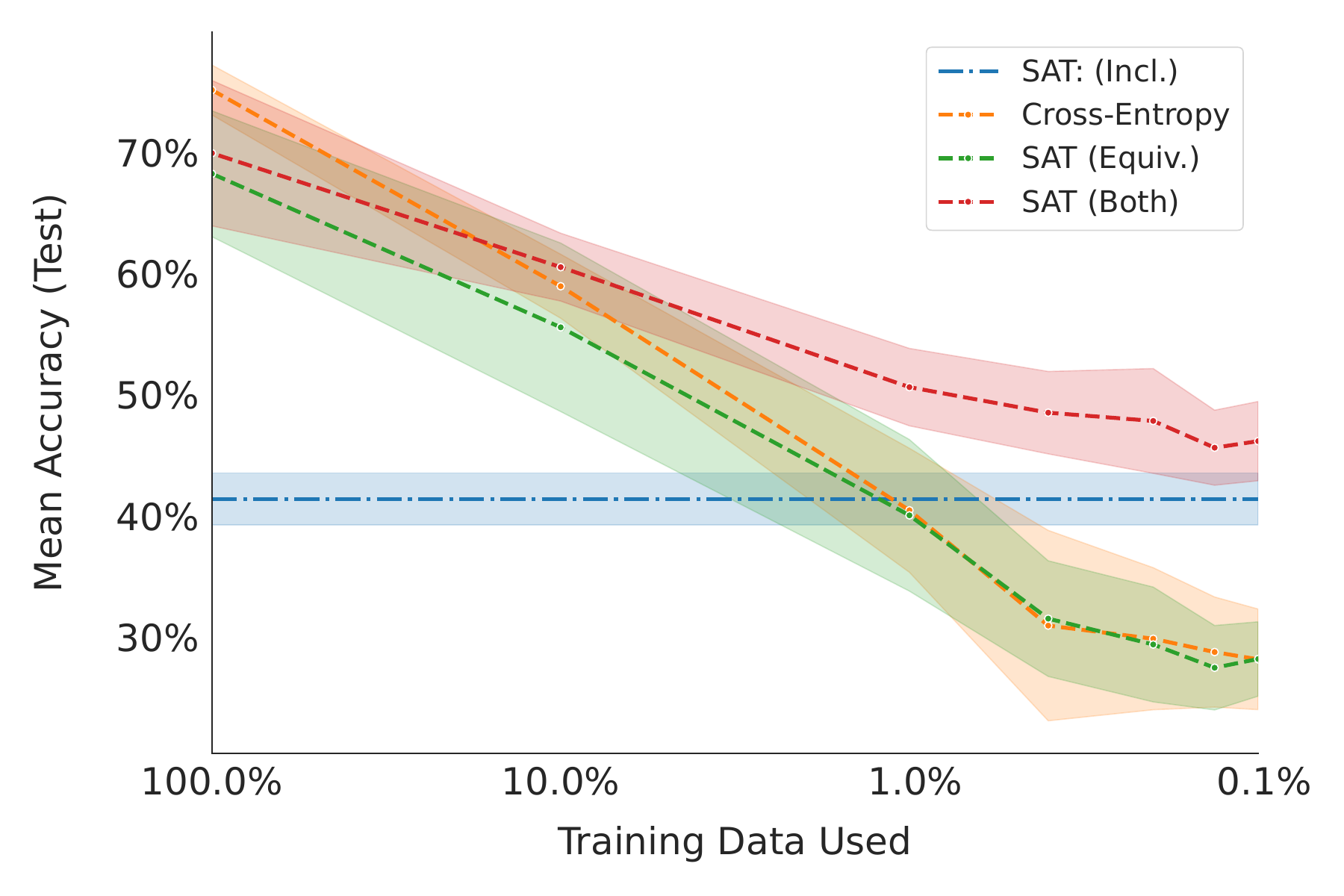}
    }
    \subfloat[Castle Island\label{fig:beach_results}]{\centering
        \includegraphics[width=0.32\textwidth]{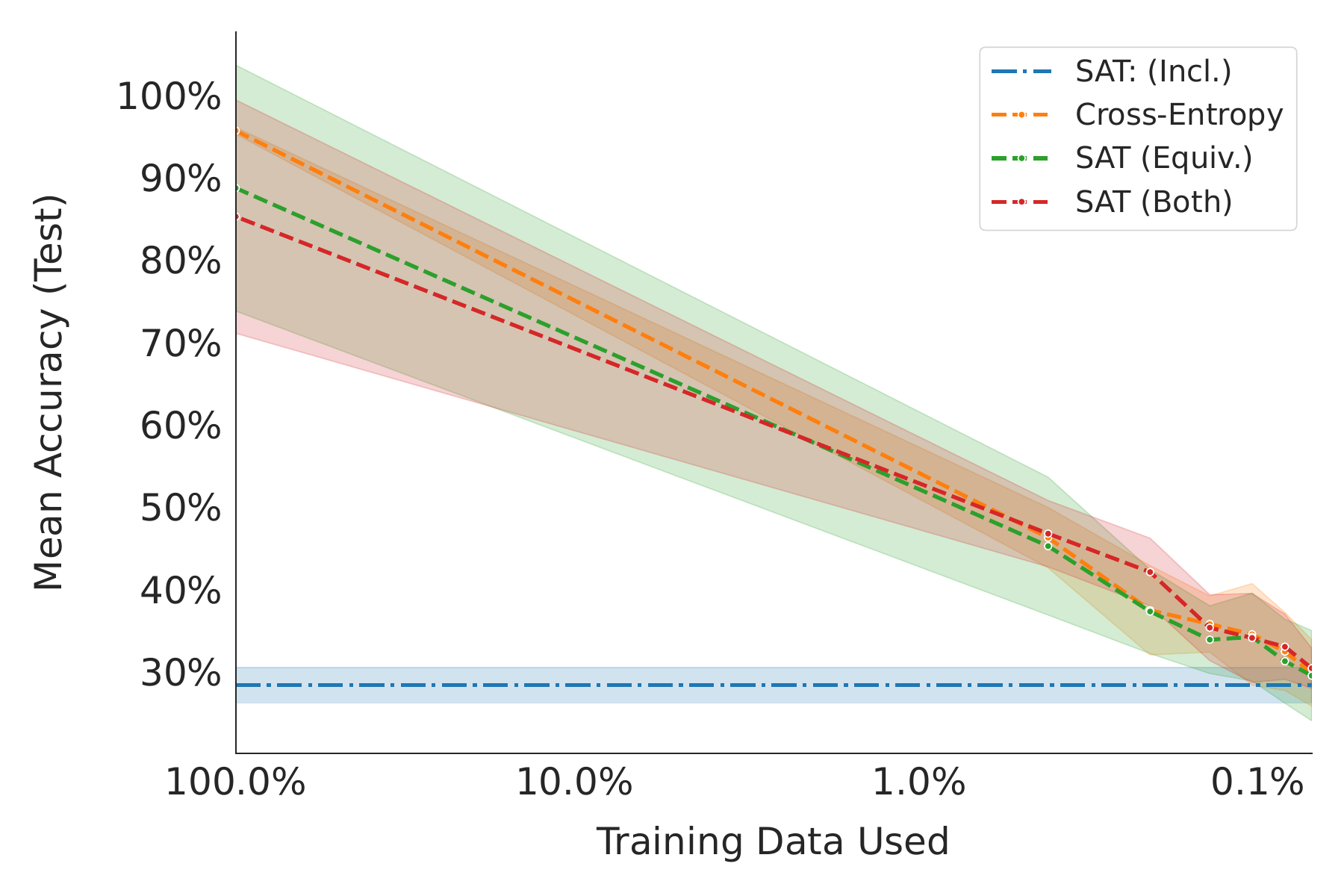}
    }
\caption{
        (a) Results of the predicate ablation for the MP3D dataset.
        (b) Results of the predicate ablation for the West Point dataset.
        (c) Results of the predicate ablation for the Castle Island dataset.
For all experiments, we run 10 trials of each loss configuration for each percentage of the training data.
        The mean of the 10 trials is shown as a dashed line, and the shaded area shows the standard deviation.
\label{fig:pred_figs}\vspace{-5mm}}
\end{figure*}

\section{Experiments}
\label{sec:experiments}
In this section, we describe experiments conducted to evaluate our approach across a variety of environments where annotated data is limited or missing completely.

\ph{Experimental Setup}
We utilize three datasets: Matterport3D (MP3D)~\cite{Chang173dv-Matterport3D},
West Point, and Castle Island.
For all datasets, we use Hydra~\cite{Hughes24ijrr-hydraFoundations} to generate 3D scene graphs.
The MP3D~\cite{Chang173dv-Matterport3D} dataset is an RGB-D dataset consisting of 90 indoor scenes.
We use the Habitat simulator~\cite{Savva19iccv-habitat} to render the visual and human-annotated semantic data for the scenes.
The provided label space has 40 semantic labels (\ie{} mpcat40~\cite{Chang173dv-Matterport3D}), of which we discard the labels \emph{gym_equipment} and \emph{beam}.
For our experiments, we use the official train, test, and validation splits of the scenes~\cite{Chang173dv-Matterport3D}.
We collect two outdoor datasets.
The first, West Point, consists of two robot trajectories where a RealSense D455 was mounted on a Clearpath Jackal rover.
These trajectories explore areas mostly consisting of roads, fields, and buildings.
The second, Castle Island, was collected from Pleasure Bay beach in Boston via a hand-held sensor payload, which consists of two trajectories exploring areas around the beach and nearby parking lot.
The trajectories from both datasets are roughly a half kilometer to a kilometer in length.
For both datasets, we use OneFormer~\cite{Jain23cvpr-OneFormer} to provide semantics (\ie low-level concepts) with the ADE20k~\cite{Zhou17cvpr-ade20k} label space.
We use LOCUS~\cite{Reinke22ral-LOCUS2} with a Velodyne mounted on the Jackal to provide trajectory estimates for the West Point dataset, and Kimera-VIO~\cite{Rosinol20icra-Kimera} for trajectory estimates of Castle Island.

\ph{Training}
We train the model using different axiom configurations that include (i) only the data-driven axioms $\phi_{Equiv}$ (label: ``SAT (Equiv.)''), (ii) only the ontology-driven axioms $\phi_{Incl}$ (label: ``SAT (Incl.)''), and (iii) both the data-driven and ontology-driven axioms (label: ``SAT (Both)'').
As a baseline, we train a GNN using cross-entropy loss (label: ``Cross-Entropy''), which is expected to be similar to the model trained with the purely data-driven axioms.
The node features for the places are the concatenation of the position and the word2vec~\cite{Mikolov13-wordRepresentation} encoding of the aggregated semantic labels of the child nodes in the mesh layer.
For the GNN, we use a graph attention network (GAT)~\cite{Velickovic18arxiv-GAT} and
tune the hyper-parameters using grid search. We use text completion (\cref{sec:text_completion}) for generating the ontology, since it generates ontologies that agree (with respect to precision) with human judgement more closely than text scoring (\cref{sec:text_scoring}); see Table~\ref{tab:onto_comparison}.
\begin{table}[t!]
    \centering
    \begin{tabular}{c c c c}
        \toprule
        Ontology & Accuracy & Precision & Recall \\
        \midrule
        Text Completion ($k{=}1$) & $0.50$ & $\mathbf{1.0}$ & $0.14$ \\
        Text Completion ($k{=}2$) & $0.52$ & $0.83$ & $0.22$ \\
        Text Completion ($k{=}3$) & $0.65$ & $\mathbf{1.0}$ & $0.40$ \\
        Text Completion ($k{=}4$) & $0.68$ & $0.85$ & $0.54$ \\
        Text Completion ($k{=}5$) & $0.71$ & $0.86$ & $0.59$ \\
        Text Scoring & $\mathbf{0.81}$ & $0.77$ & $\mathbf{0.95}$ \\
        \bottomrule
    \end{tabular}
        \caption{Evaluation of ontologies produced by text scoring and text completion against an ontology created using human judgement for the MP3D label space with a temperature $K{=}10$ for text scoring. Best in bold.}\label{tab:onto_comparison}
     \vspace{-3mm}
\end{table}
 For the connectives and quantifiers, we use a product configuration except for the implication operator where we use the Goguen operator~\cite{Van22ai-diffFuzzyLogic}.
We use stochastic gradient descent with the Adam optimizer, and we run the optimization until the loss has converged (\eg{} changes less than 1e-06 for 10 epochs) or 1000 epochs is reached.
The model with the best validation accuracy is retained for testing.
Each model is trained 10 times using a random initialization of the model parameters, and the statistics are reported in the results.
We implement the architectures using PyTorch Geometric~\cite{Fey19iclrwk-pytorchGeometric} and LTNTorch~\cite{Badreddine22ai-LTN}.

\ph{Results}
We conduct an ablation study comparing the cross-entropy baseline and the different axioms configurations for the LTN.
The accuracy (\ie percentage of correctly classified places) for all datasets is reported in~\cref{fig:pred_figs}, and the average runtime for testing per scene graph on a CPU (Intel i7-11800 at 2.3GHz) is 0.027s for MP3D, 0.067s for West Point, and 0.207s for Castle Island.
For each dataset, we evaluate the performance for $100\%$, $10\%$, $1\%$, $0.4\%$, $0.2\%$, and $0.1\%$ of the training data.
We use the train-test-validation splits for MP3D as described in~\cite{Chang173dv-Matterport3D}, and we mask these percentages of places from the training scene to simulate a scenario where annotated data is limited.
For West Point and Castle Island, we train in a semi-supervised manner where we combine both graphs ($70\%$ training, $15\%$ validation and testing), as the datasets are too small to generalize from a single training scene to a single testing scene.

\begin{figure*}
    \centering
    \subfloat[MP3D Mesh]{\centering
\includegraphics[width=0.265\textwidth,trim={8.3cm 0mm 7.2cm 0mm},clip]{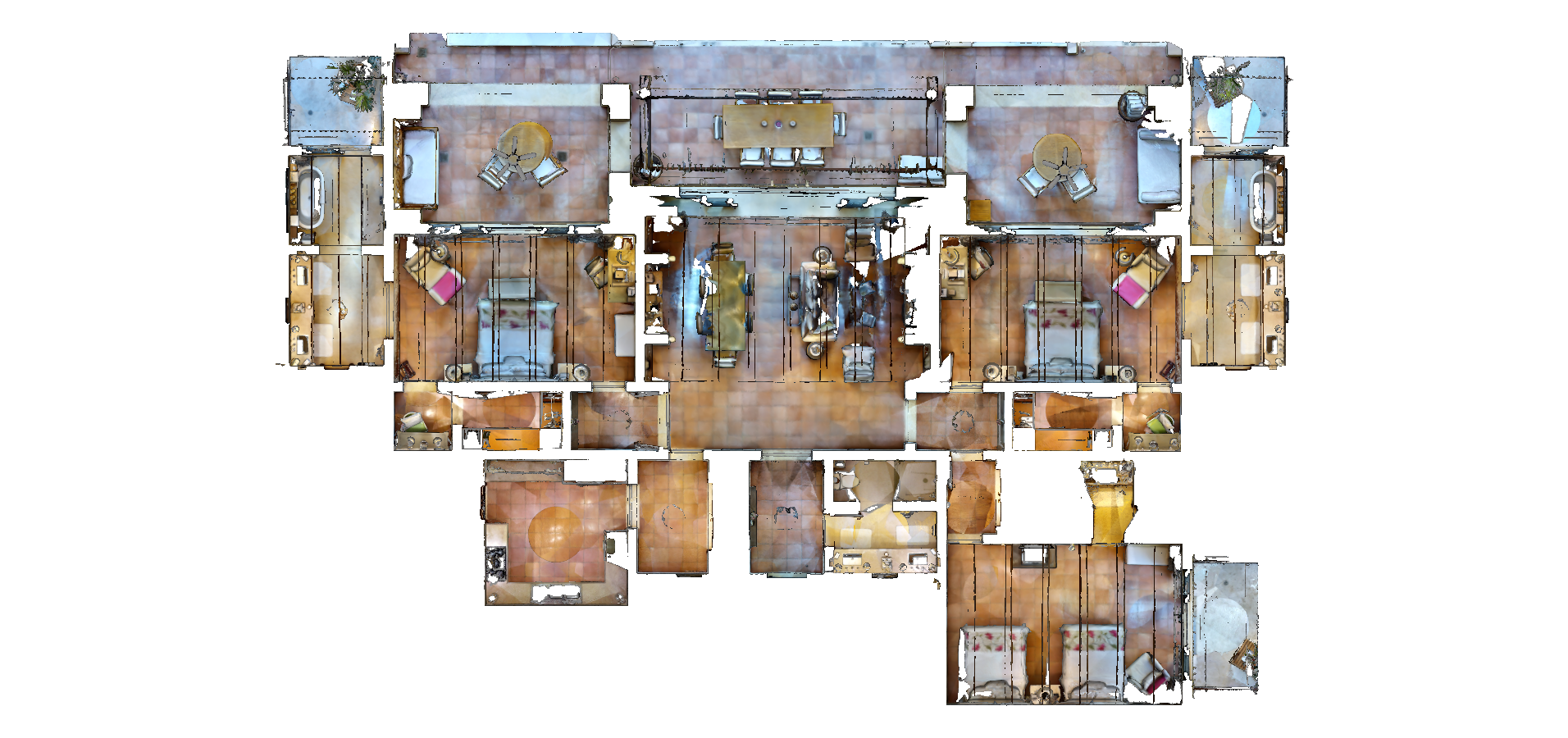}
    }
    \subfloat[MP3D\label{fig:mp3d_qualitative}]{\centering
\includegraphics[width=0.295\textwidth,trim={2.3cm 0mm 7.2cm 0mm},clip]{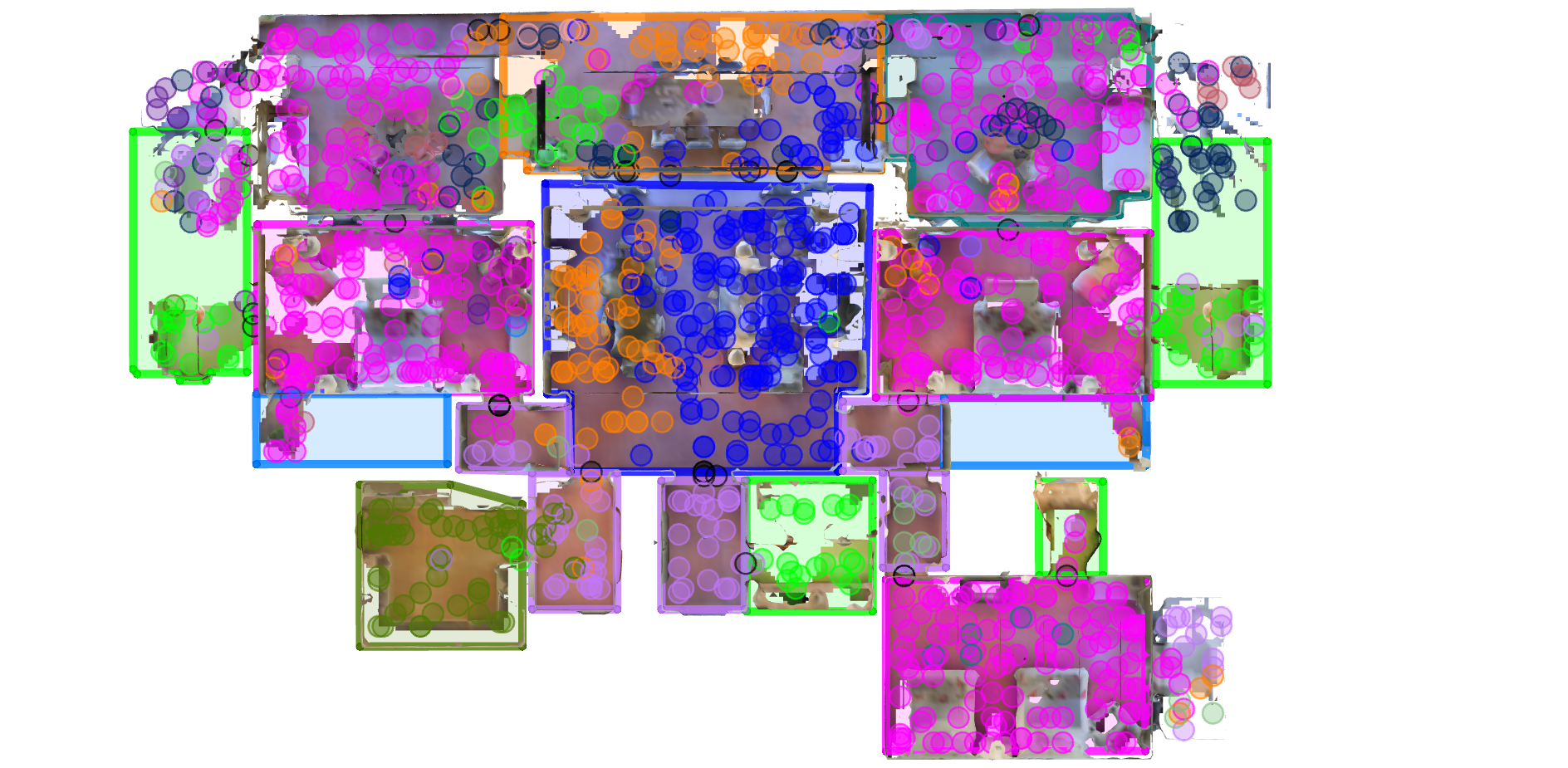}
    }
    \subfloat[MP3D Color Legend]{\centering
        \raisebox{5mm}{\includegraphics[width=0.44\textwidth,trim={0 0 2cm 0},clip]{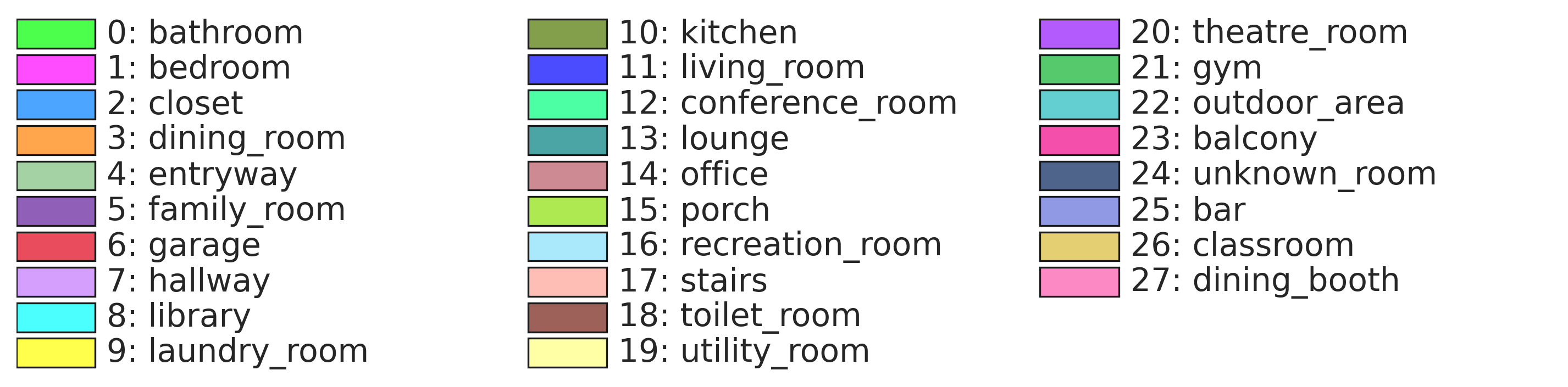}}
    } \\
    \subfloat[West Point\label{fig:westpoint_qualitative}]{\centering
\includegraphics[width=0.34\textwidth,trim={0mm 0mm 0mm 0mm},clip]{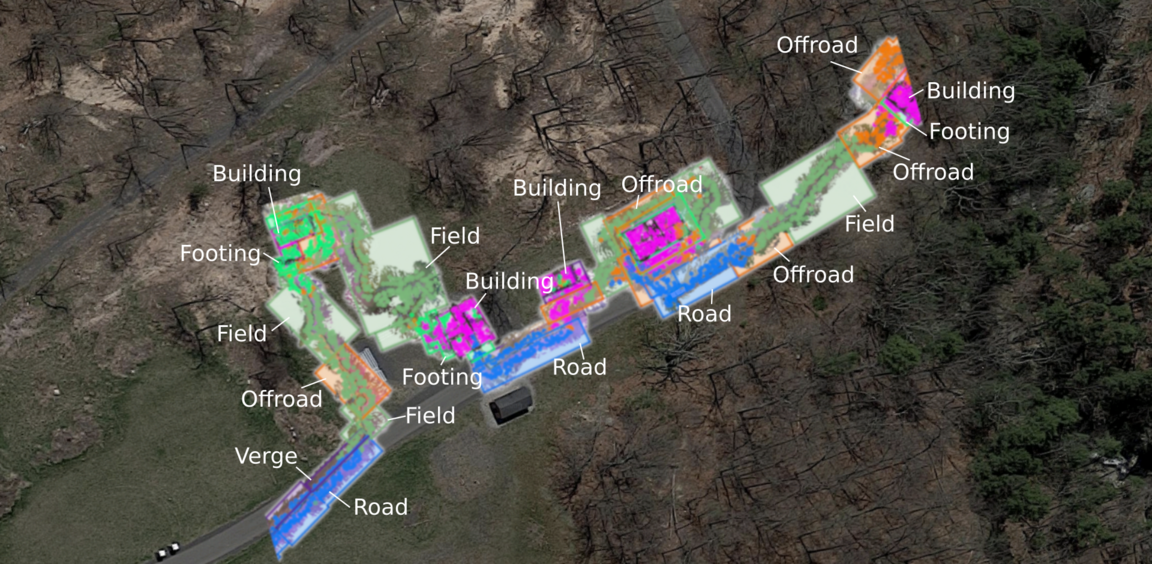}
    }
    \subfloat[Castle Island\label{fig:beach_qualitative}]{\centering
\includegraphics[width=0.35\textwidth,trim={0cm 0mm 0cm 0mm},clip]{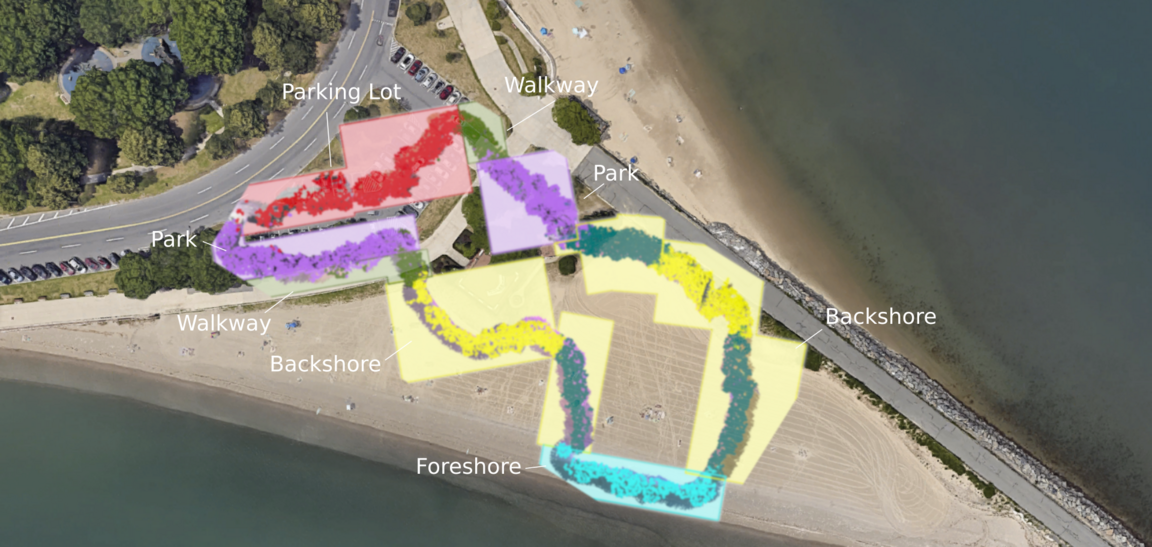}
    }
    \subfloat[Outdoor Color Legend]{\centering
        \raisebox{5mm}{\includegraphics[width=0.31\textwidth,trim={0 0 2.5cm 0},clip]{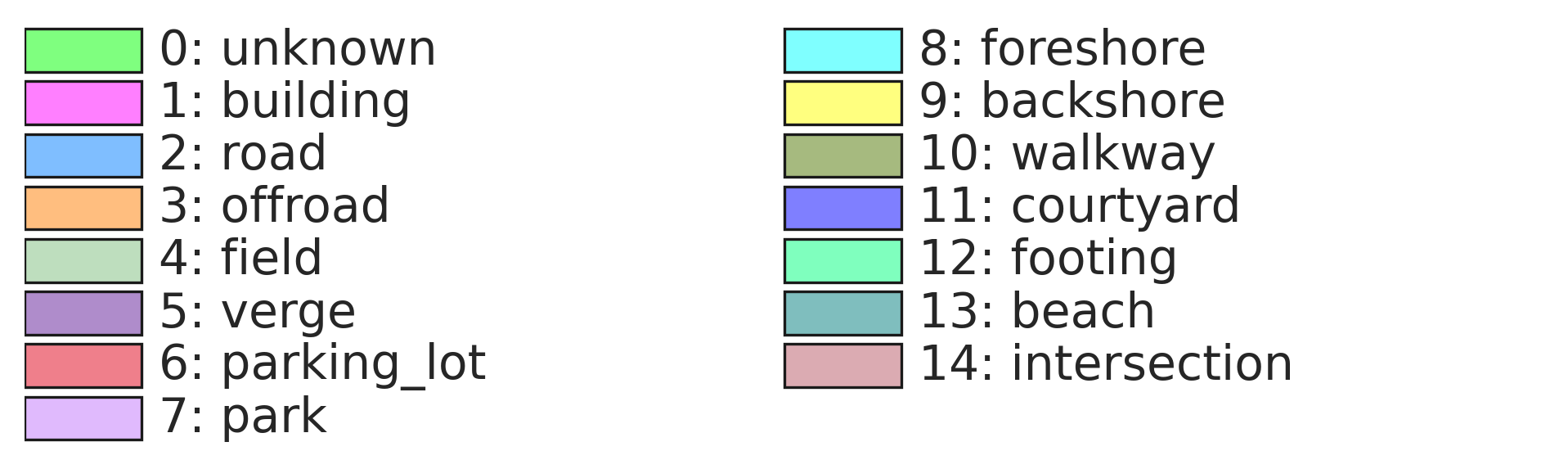}}
    }
    \caption{(a) The ``q9vSo1VnCiC'' scene from MP3D and (b) classified place nodes for the ``q9vSo1VnCiC'' scene along with (c) associated color legend for node labels.
        (d) Classified place nodes for one of the West Point trajectories.
        (e) Classified place nodes for on of the Castle Island trajectories.
        (f) Associated color legend for both West Point and Castle Island.
        The ground truth regions are drawn as shaded polygons, while the place nodes are all drawn as circles.
    }\label{fig:qualitative_dsgs}
    \vspace{-5mm}
\end{figure*}

The data-driven approaches (our approach, the GNN baseline, and our data-driven axiom variant) show similar performance when trained on the full MP3D test dataset; see~\cref{fig:mp3d_results}.
However, as less data is used during training, the purely data-driven approaches tend towards performance comparable with randomly guessing labels, while the proposed approach does not fall below the purely ontology-driven approach.
The same trend is visible in~\cref{fig:westpoint_results} for the West Point dataset.
The data-driven approaches in~\cref{fig:beach_results} for the Castle Island dataset do not have a significant gap in performance compared to the ontology-driven approaches.
This is due to a lack in distinguishing low-level concepts for the high-level concepts in the scene.
\begin{table}[t!]
    \centering
    \begin{tabular}{c c c c}
        \toprule
        Ontology & MP3D & West Point & Castle Island \\
        \midrule
        $k=1$ & $14.0 \pm  2.0$ & $37.8 \pm  0.3$ & $16.1 \pm  3.6$ \\
        $k=2$ & $16.7 \pm  3.6$ & $11.9 \pm 14.7$ & $\mathbf{28.7} \pm  \mathbf{2.2}$ \\
        $k=3$ & $23.1 \pm  0.2$ & $\mathbf{42.7} \pm \mathbf{ 0.2}$ & $24.5 \pm  0.6$ \\
        $k=4$ & $ 2.7 \pm  0.0$ & $39.9 \pm  4.7$ & $28.7 \pm 8.5$ \\
        $k=5$ & $12.3 \pm  0.0$ & $38.8 \pm 12.0$ & $19.7 \pm  7.1$ \\
        Custom & $\mathbf{27.7} \pm \mathbf{ 0.4}$ & $27.7 \pm  1.3$ & $22.7 \pm  1.8$ \\
        \bottomrule
    \end{tabular}
        \caption{Comparison of mean accuracy achieved by the ``SAT (Incl.)'' loss configuration across the three datasets between the LLM generated ontologies ($k=1, \ldots, 5$) versus a hand-generated ontology for the same low-level label space and high-level concepts. Best in bold.}\label{tab:ontology}
     \vspace{-3mm}
\end{table}
 \begin{table}[t!]
    \centering
    \begin{tabular}{c @{\hspace{1.0\tabcolsep}} c @{\hspace{1.0\tabcolsep}} c @{\hspace{1.0\tabcolsep}} c @{\hspace{1.0\tabcolsep}} c}
        \toprule
        & \multicolumn{4}{c}{\% of Data Kept for MP3D w/ Partial Labels } \\
        \cmidrule{2-5}
        Loss & 100\% & 0.4\% & 0.2\% & 0.1\% \\
        \midrule
        CrossEntropy     & $24.1 \pm 2.4$                   & $16.7 \pm 5.6$          & $12.7 \pm  4.7$ & $12.5 \pm 5.1$ \\
        SAT (Equiv.) & $24.7 \pm 1.4$                   & $15.8 \pm 4.0$          & $12.8 \pm  5.5$ & $11.7 \pm 5.1$ \\
        SAT (Incl.)   & $22.7 \pm 0.6$                   & $\mathbf{22.7} \pm \mathbf{0.6}$ & $22.7 \pm 0.6$ & $22.7 \pm  0.6$ \\
        SAT (Both)        & $\mathbf{32.5} \pm \mathbf{1.2}$ & $20.6 \pm 1.4$          & $\mathbf{22.8} \pm \mathbf{1.9}$ & $\mathbf{23.3} \pm  \mathbf{1.4}$ \\
        \bottomrule
    \end{tabular}
        \caption{Accuracy for different holdout percentages of training data with labels masked for \emph{bathroom}, \emph{kitchen}, \emph{hallway}, \emph{bedroom}, \emph{living room}, and \emph{family room}. No data held out during testing. Best in bold.}\label{tab:mask}
     \vspace{-5mm}
\end{table}
 We choose $k$ by generating ontologies with multiple values ($k{=}\{1,2,3,4,5\}$), and the $k$ with the highest test accuracy is used in the ablations; alternatively, $k$ may be chosen by evaluating against a smaller ontology created based on human judgement.
We compare these generated ontologies with a hand-generated ontology and report the results in Table~\ref{tab:ontology}.

Finally, a 3D scene graph obtained by our method for each dataset is shown in~\cref{fig:qualitative_dsgs}.
In general, the high-level concepts generated by our approach are plausible for the associated scenes.
For certain scenes, our approach struggles to distinguish similar high-level concepts.
This is apparent for the beach (\eg ground truth labels ``backshore'' are classified as ``beach'' instead).
Additionally, low-level concepts that contradict the spatial ontology cause our model to misclassify certain regions.
This is clear in the MP3D scene shown in~\cref{fig:mp3d_qualitative}; the lounge in the top-right of the building is misclassified as a bedroom due to the presence of a day-bed in the room, which is labeled as ``bed'' in the ground truth semantics of the MP3D dataset.
Lastly, we show in the MP3D dataset that the ontology-driven models allow predicting labels unseen at training time.
This scenario is simulated by masking all place nodes for the following set of labels: \emph{bathroom}, \emph{kitchen}, \emph{hallway}, \emph{bedroom}, \emph{living room}, and \emph{family room}.
In this scenario, the ontology-driven models significantly outperform the purely data-driven models as shown in~\cref{tab:mask} since labels for the removed classes are missing during training.

\section{Conclusion}
We propose the first approach to build hierarchical 3D scene graphs of arbitrary (indoor and outdoor) environments.
Towards this goal, we leverage Large Language Models to build a \emph{spatial ontology} that defines the set of concepts and relations relevant for robot operation.
We show how to use such a spatial ontology for 3D scene graph construction
using \emph{Logic Tensor Networks} (LTN), which allow adding logical rules, or \emph{axioms} (\eg ``a beach contains sand''), during training and prediction.
We test our approach in a variety of datasets, including indoor, rural, and coastal environments, and show that it leads to a significant increase in the quality of the 3D scene graph generation with sparsely annotated data, and it produces reasonable results even in a zero-shot setting.
This work opens several avenues for future research, including constructing 3D scene graphs with multiple high-level layers, incorporating different types of relations in the ontology, and more powerful feature representations.

\begingroup
{\scriptsize \bibliographystyle{IEEEtran}

}
\endgroup

\onecolumn
\twocolumn
\appendix

\subsection{Details for Training}
Here, we provide additional details on training the proposed model.

\ph{Implementation}
We implement the learning architectures using PyTorch Geometric~\cite{Fey19iclrwk-pytorchGeometric} version 2.3.1 built with CUDA verion 11.7 and LTNTorch~\cite{Badreddine22ai-LTN} version 1.0.1, and we use GPT-4 for generating the spatial ontology.
For the GNN, we use GAT~\cite{Velickovic18iclr-GAT} for the message passing layers, and we use ReLU activation functions and dropout between message passing layers where dropout is enabled for training but disabled for testing.
After the messaging passing layers in the GNN, the hidden states are passed to the MLP with a linear layer followed by ReLU activation functions followed by another linear layer, then a softmax layer is applied for predicting the node labels.
The input and hidden dimensions for the MLP are the same as the hidden dimensions for the message passing layers in the GNN.

\ph{Hyper-Parameter Tuning}
We provide details on the method applied to tune the hyper-parameters of our model.
Specifically, we search over the following sets:
\begin{itemize}
    \item Message passing iterations: $[1,2,3]$
    \item Message passing hidden dimensions: $[16,32,64]$
    \item Number of attention heads: $[1,2,4]$
    \item Learning rate: $[0.01, 0.001, 0.0001]$
    \item Dropout probability: $[0.1, 0.25, 0.5]$
    \item Weight decay: $[0.005, 0.0005, 0.00005]$

\end{itemize}

In order to provide fair comparison between the different axioms configurations, a grid search was performed over these hyper-parameters for the Cross-Entropy configuration as well as each of the LTN configurations.
For the Cross-Entropy configuration and the LTN configurations, the number of convolutional layers, the learning rate, and the weight decay made the most impact on the accuracy of the model (with weight decay making less impact than convolutional layers and learning rate).
Thus, we tune the hyper-parameters for the message passing iterations while keeping the remaining parameters constant at the median values.
The learning rate and weight decay are next tuned by searching over all combinations of these values, and the grid search is performed over the dropout rate and the number of attention heads.
The grid search described above was performed with $1000$ epochs using the MP3D dataset, and we repeated this process for the learning rate and weight decay for both the West Point and Castle Island datasets, and optimal values of hyper-parameters remained the same as for the MP3D dataset.
The optimal hyper-parameters based on the grid search are the same for each of the Cross-Entropy and different axioms configuration; see Table~\ref{tab:params} for the selected values of the hyper-parameters.

\begin{table}[h!]
\centering
    \begin{tabular}{c cc}
        \toprule
        Layers                & 3\\
        Hidden Dimension      & 32\\
        Learning Rate         & 0.001\\
        Dropout               & 0.25\\
        $L_2$ Regularization  & 5e-05\\
        Attention Heads       & 4\\
        \bottomrule
    \end{tabular}
    \caption{Hyper-parameters for training the proposed model where the same hyper-parameters are used for the Cross-Entropy configuration and the LTN configurations.}\label{tab:params}
    \vspace{-4mm}
\end{table}

Additionally, we must select the fuzzy logical operators used for the axioms in the LTN along with the hyper-parameters associated with these operators.
For the implication operator, we use the Goguen operator instead of the more commonly used Reichenbach operator, as the Goguen operator provided better accuracy.
Additionally, we use the $p$-mean error for the aggregating operator $\texttt{SatAgg}(\cdot)$ as described in~\cite{Badreddine22ai-LTN}, and we use $p{=}2$ for the data-driven axioms (\ie equivalence axioms) and $p{=}4$ for the ontology-driven axioms (\ie inclusive axioms).
The model was trained with $p = [2,4]$ for each axiom configuration, and the optimal value was selected; however, the difference in accuracy was insignificant.

Additionally, for the experiments where only a percentage of annotations for the data is kept during training, the annotations kept are selected randomly; furthermore, we seed the random number generator identically between the experiments for the Cross-Entropy configuration and the different LTN configurations, so the results are consistent and reproducible between experiments.
Furthermore, we use a different seed for each of the Monte-Carlo trials in the experiments (\eg trial $1$ is seeded as $1$, trial $2$ is seeded as $2$, and so on for the Cross-Entropy and LTN configurations).

\ph{Runtime}
The model is trained using an AMD Ryzen Threadripper PRO 5955WX for the CPU and an NVIDIA GeForce RTX 4090 for the GPU.
The runtime for training and testing the proposed model is provided in Table~\ref{tab:runtime}.
Note, the number of parameters of the model does not change between the Cross-Entropy and LTN configurations; thus, the runtime during testing is not affected by the configuration used during training.

\begin{table}[h!]
    \centering
    \begin{tabular}{c c c c}
        \toprule
        Loss & MP3D & West Point & Castle Island \\
        \midrule
        Cross-Entropy  & $0.018$s & $0.002$s & $0.005$s \\
        SAT (Equiv.)   & $0.035$s & $0.003$s & $0.007$s \\
        SAT (Incl.)    & $0.027$s & $0.003$s & $0.006$s \\
        SAT (Both.)    & $0.038$s & $0.003$s & $0.008$s \\
        \midrule
        Testing (CPU): & $0.027s$     & $0.067$s   & $0.207$s \\
\bottomrule
    \end{tabular}
        \caption{Comparison of runtime for the proposed model for training and testing. The reported time is averaged across all epochs and all graphs for training and across all graphs for testing. The average number of (nodes, edges) for the graphs in each dataset are $(1283,7407)$ for MP3D, $(3930,15057)$ for West Point, and $(11059,69784)$ for Castle Island.}\label{tab:runtime}
     \vspace{-4mm}
\end{table}

\subsection{Evaluation of Spatial Ontology}

Here, we provide additional details regarding the evaluation of the generated ontologies using text scoring and text completion against a ground truth ontology based on human judgement.
Specifically, we annotated a random sample of relations between low-level and high-level concepts (\eg a "sink" is found in a "kitchen") as being "likely", "sometimes", and "rarely".
We sample 38 pairs where an edge exists between 22 of the pairs (annotated as "likely") and an edge does not exist between 16 of the pairs (annotated as "sometimes" or "rarely").
Next, we compute the accuracy, precision, and recall for the spatial ontologies generated by the text scoring approach (using a temperature of $K=10$) and the text completion approach (using $k=\{1,2,3,4,5\}$) against the human annotated relations; see Table~\ref{tab:onto_comparison}.
For the MP3D dataset, there are 31 high-level and 39 low-level concepts, and for the outdoor datasets, there are 15 high-level and 150 low-level concepts.
For the text completion approach, the number of edges is $k|\calH|$, which is $78$ ($k=3$) for MP3D, $42$ for West Point ($k=3$), and $28$ for Castle Island ($k=2$).
Note, we do not include rooms labeled unknown when generating the spatial ontology.
The number of edges is 99 for the MP3D dataset using the text scoring approach, and the text scoring approach was not applied to the outdoor datasets.

While the accuracy and recall for the text scoring approach significantly outperforms the text completion approach, the precision is significantly better for the text completion approach.
Since the ontology is generally sparse, the precision of the spatial ontology is a better performance indicator for the LTN, which is consistent with the results.
Recall, the value of $k{=}3$ provides the best accuracy for the LTN for the MP3D dataset as shown in Table~\ref{tab:ontology}, which is also the spatial ontology tied with $k{=}1$ for the highest precision (where $k{=}3$ has higher accuracy and recall than $k{=}1$).
Furthermore, this can be observed by comparing the accuracy for grounding the high-level concepts in 3D scene graphs using the text completion versus the text scoring approaches; see Table~\ref{tab:onto_ltn_comparison}.

In contrast to the text scoring approach, the text completion approach requires choosing a value of $k$, which restricts the number of edges incident to each low-level concept.
The number of edges incident to each low-level concept in the text scoring approach varies between low-level concepts, which could be preferrable; however, many incident edges to a particular low-level concept indicates the concept may not be useful for distinguishing high-level concepts (\eg "wall" is common to most rooms in a building making it a poor indicator for a room type).

\begin{table}[h!]
    \centering
    \begin{tabular}{c @{\hspace{1.0\tabcolsep}} c @{\hspace{1.0\tabcolsep}} c @{\hspace{1.0\tabcolsep}} c @{\hspace{1.0\tabcolsep}} c}
        \toprule
& \multicolumn{2}{c}{SAT (Incl.)} & \multicolumn{2}{c}{SAT (Both)} \\
        \cmidrule(l{2pt}r{2pt}){2-3} \cmidrule(l{2pt}r{2pt}){4-5}
        \% Kept & Text Scoring & Text Completion & Text Scoring & Text Completion \\
        \midrule
        100\% & $12.1 \pm 4.6$ & $\mathbf{23.1} \pm \mathbf{0.2}$ & $52.7 \pm 4.6$  & $\mathbf{52.9} \pm \mathbf{1.8}$ \\
        10\%  & $12.1 \pm 4.6$ & $\mathbf{23.1} \pm \mathbf{0.2}$ & $50.7 \pm 1.6$  & $\mathbf{53.2} \pm \mathbf{1.1}$ \\
        1\%   & $12.1 \pm 4.6$ & $\mathbf{23.1} \pm \mathbf{0.2}$ & $27.3 \pm 2.4$  & $\mathbf{35.9} \pm \mathbf{3.3}$ \\
        0.4\% & $12.1 \pm 4.6$ & $\mathbf{23.1} \pm \mathbf{0.2}$ & $19.8 \pm 2.8$  & $\mathbf{29.8} \pm \mathbf{3.2}$ \\
        0.2\% & $12.1 \pm 4.6$ & $\mathbf{23.1} \pm \mathbf{0.2}$ & $18.0 \pm 1.1$  & $\mathbf{26.8} \pm \mathbf{2.3}$ \\
        0.1\% & $12.1 \pm 4.6$ & $\mathbf{23.1} \pm \mathbf{0.2}$ & $16.7 \pm 1.0$  & $\mathbf{25.1} \pm \mathbf{2.2}$ \\
        \bottomrule
    \end{tabular}
        \caption{Comparison of accuracy of the proposed model between the "text scoring" and "text completion" methods for generating the spatial ontology for different percentages of data kept during training. Note, the predicates for SAT (Incl.) do not use labeled data; thus, the performance is identical across data kept during training. No data held out during testing. Best in bold.}\label{tab:onto_ltn_comparison}
\end{table}

\end{document}